\documentclass[letterpaper,english]{article}
\usepackage[T1]{fontenc}
\usepackage[latin1]{inputenc}
\usepackage{babel}
\usepackage{graphics}
\usepackage{algorithm}
\usepackage{subfigure}

\makeatletter

\providecommand{\LyX}{L\kern-.1667em\lower.25em\hbox{Y}\kern-.125emX\@}

 \usepackage{verbatim}

\usepackage{naacl2001}
\usepackage{algorithm}
\usepackage{epsfig}
\usepackage{subfigure}

\usepackage{amstex}

\usepackage{amstex}
\makeatother
\begin{document}

\title{Transformation-Based Learning in the Fast Lane}

\author{Grace Ngai\( ^{\dagger ,\ddagger } \) \and Radu Florian\( ^{\dagger } \)\\
\{gyn,rflorian\}@@cs.jhu.edu\\
\vspace*{0.1cm}\\
\begin{tabular}{cc}
\( ^{\dagger } \)\parbox[t]{5cm}{\centering Johns Hopkins University\\Baltimore, MD 21218,~USA}&
\( ^{\ddagger } \)\parbox[t]{4.3cm}{\centering Weniwen Technologies\\ Hong Kong}\\
\end{tabular}}

\maketitle
\begin{abstract}
Transformation-based learning has been successfully employed to solve
many natural language processing problems. It achieves state-of-the-art
performance on many natural language processing tasks and does not
overtrain easily. However, it does have a serious drawback: the training
time is often intorelably long, especially on the large corpora which
are often used in NLP. In this paper, we present a novel and realistic
method for speeding up the training time of a transformation-based
learner without sacrificing performance. The paper compares and contrasts
the training time needed and performance achieved by our modified
learner with two other systems: a standard transformation-based learner,
and the ICA system \cite{hepple00:tbl}. The results of these experiments
show that our system is able to achieve a significant improvement
in training time while still achieving the same performance as a standard
transformation-based learner. This is a valuable contribution to systems
and algorithms which utilize transformation-based learning at any
part of the execution.
\end{abstract}

\section{Introduction}

\label{section:introduction}

Much research in natural language processing has gone into the development
of rule-based machine learning algorithms. These algorithms are attractive
because they often capture the linguistic features of a corpus in
a small and concise set of rules.

Transformation-based learning (TBL) \cite{brill95:tagging} is one
of the most successful rule-based machine learning algorithms. It
is a flexible method which is easily extended to various tasks and
domains, and it has been applied to a wide variety of NLP tasks, including
part of speech tagging \cite{brill95:tagging}, noun phrase chunking
\cite{ramshaw99:basenp}, parsing \cite{brill96:transformation_parsing},
phrase chunking \cite{florian00:tbldt}, spelling correction \cite{mangu97:cssc},
prepositional phrase attachment \cite{brill94:pp-attach}, dialog
act tagging \cite{samuel98:dialogact}, segmentation and message understanding
\cite{day97:alembic}. Furthermore, transformation-based learning
achieves state-of-the-art performance on several tasks, and is fairly
resistant to overtraining \cite{ramshaw94}.

Despite its attractive features as a machine learning algorithm, TBL
does have a serious drawback in its lengthy training time, especially
on the larger-sized corpora often used in NLP tasks. For example,
a well-implemented transformation-based part-of-speech tagger will
typically take over 38 hours to finish training on a 1 million word
corpus. This disadvantage is further exacerbated when the transformation-based
learner is used as the base learner in learning algorithms such as
boosting or active learning, both of which require multiple iterations
of estimation and application of the base learner. In this paper,
we present a novel method which enables a transformation-based learner
to reduce its training time dramatically while still retaining all
of its learning power. In addition, we will show that our method scales
better with training data size. 
 
\section{Transformation-based Learning}

\label{section:tbl}

The central idea of transformation-based learning (TBL) is to learn
an ordered list of rules which progressively improve upon the current
state of the training set. An initial assignment is made based on
simple statistics, and then rules are greedily learned to correct
the mistakes, until no net improvement can be made.

The following definitions and notations will be used throughout the
paper: 

\begin{itemize}
\item \vspace*{-1mm}The \textit{sample space} is denoted by \( \mathcal{S} \);
\item \vspace*{-1mm}\( \mathcal{C} \) denotes the set of possible \textit{classifications}
of the samples; 
\item \vspace*{-1mm}\( C[s] \) denotes the classification associated with
a sample \( s \), and \( T[s] \) denotes the true classification
of \( s \);
\item \vspace*{-1mm}\( p \) will usually denote a \textit{predicate} defined
on \( \mathcal{S} \); 
\item \vspace*{-1mm}A \textit{rule} \( r \) is defined as a predicate \,-\,
class label pair, \( (p,t) \), where \( t\in \mathcal{C} \) is called
the \textit{target} of \( r \); 
\item \vspace*{-1mm}\( \mathcal{R} \) denotes the set of all rules; 
\item \vspace*{-1mm}If \( r=(p,t) \), \( p_{r} \) will denote \( p \)
and \( t_{r} \) will denote \( t \); 
\item \vspace*{-1mm}A rule \( r=(p_{r},t_{r}) \) \textit{applies} to a
sample \( s \) if \( p_{r}(s)=true \) and \( t_{r}\neq C[s] \);
the resulting sample is denoted by \( r(s) \). 
\end{itemize}
\noindent Using the TBL framework to solve a problem assumes the existence
of: 

\begin{itemize}
\item \vspace*{-1mm}An initial class assignment. This can be as simple as
the most common class label in the training set, or it can be the
output of another classifier. 
\item \vspace*{-1mm}A set of allowable templates for rules. These templates
determine the types of predicates the rules will test; they have the
largest impact on the behavior of the system. 
\item \vspace*{-1mm}An objective function \( f \) for learning. Unlike
in many other learning algorithms, the objective function for TBL
will directly optimize the evaluation function. A typical example
is the difference in performance resulting from applying the rule:\[
f\left( r\right) =good\left( r\right) -bad\left( r\right) \]
where \[
\begin{array}{c}
good\left( r\right) =\left| \left\{ s|C\left[ s\right] \neq T\left[ s\right] \wedge C\left[ r\left( s\right) \right] =T\left[ s\right] \right\} \right| \\
bad\left( r\right) =\left| \left\{ s|C\left[ s\right] =T\left[ s\right] \wedge C\left[ r\left( s\right) \right] \neq T\left[ s\right] \right\} \right| 
\end{array}\]

\end{itemize}
Since we are not interested in rules that have a negative objective
function value, only the rules that have a positive \( good\left( r\right)  \)
need be examined. This leads to the following approach:

\begin{enumerate}
\item \vspace*{-1mm}Generate the rules (using the rule template set) that
correct at least an error (i.e. \( good\left( r\right) >0 \)), by
examining all the incorrect samples (\( s \) s.t. \( C\left[ s\right] \neq T\left[ s\right]  \)); 
\item \vspace*{-1mm}Compute the values \( bad\left( \cdot \right)  \) for
each rule \( r \) such that \( good(r)>f(b) \) , storing at each
point in time the rule \( b \) that has the highest score; while
computing \( bad(r) \), skip to the next rule when \[
f\left( r\right) <f\left( b\right) \]

\end{enumerate}
\vspace*{-2mm}The system thus learns a list of rules in a greedy fashion,
according to the objective function. When no rule that improves the
current state of the training set beyond a pre-set threshold can be
found, the training phase ends. During the application phase, the
evaluation set is initialized with the initial class assignment. The
rules are then applied sequentially to the evaluation set in the order
they were learned. The final classification is the one attained when
all rules have been applied.

\subsection{Previous Work}

As was described in the introductory section, the long training time
of TBL poses a serious problem. Various methods have been investigated
towards ameliorating this problem, and the following subsections detail
two of the approaches.

\subsubsection{The Ramshaw \& Marcus Approach}

One of the most time-consuming steps in transformation-based learning
is the updating step. The iterative nature of the algorithm requires
that each newly selected rule be applied to the corpus, and the current
state of the corpus updated before the next rule is learned.

Ramshaw \& Marcus \shortcite{ramshaw94} attempted to reduce the training
time of the algorithm by making the update process more efficient.
Their method requires each rule to store a list of pointers to samples
that it applies to, and for each sample to keep a list of pointers
to rules that apply to it. Given these two sets of lists, the system
can then easily: 

\begin{enumerate}
\item identify the positions where the best rule applies in the corpus;
and 
\item update the scores of all the rules which are affected by a state change
in the corpus. 
\end{enumerate}
These two processes are performed multiple times during the update
process, and the modification results in a significant reduction in
running time.

The disadvantage of this method consists in the system having an unrealistically
high memory requirement. For example, a transformation-based text
chunker training upon a modestly-sized corpus of 200,000 words has
approximately 2 million rules active at each iteration. The \textit{additional}
memory space required to store the lists of pointers associated with
these rules is about 450 MB, which is a rather large requirement to
add to a system.\footnote{%
We need to note that the 200k-word corpus used in this experiment
is considered small by NLP standards. Many of the available corpora
contain over 1 million words. As the size of the corpus increases,
so does the number of rules and the additional memory space required.
}

\subsubsection{The ICA Approach}

The ICA system \cite{hepple00:tbl} aims to reduce the training time
by introducing independence assumptions on the training samples that
dramatically reduce the training time with the possible downside of
sacrificing performance.

To achieve the speedup, the ICA system disallows any interaction between
the learned rules, by enforcing the following two assumptions: 

\begin{itemize}
\item \textit{Sample Independence} \,---\, a state change in a sample (e.g.
a change in the current part-of-speech tag of a word) does not change
the context of surrounding samples. This is certainly the case in
tasks such as prepositional phrase attachment, where samples are mutually
independent. Even for tasks such as part-of-speech tagging where intuition
suggests it does not hold, it may still be a reasonable assumption
to make if the rules apply infrequently and sparsely enough. 
\item \textit{Rule Commitment} \,---\, there will be at most one state change
per sample. In other words, at most one rule is allowed to apply to
each sample. This mode of application is similar to that of a decision
list \cite{rivest87:decision_lists}, where an sample is modified
by the first rule that applies to it, and not modified again thereafter.
In general, this assumption will hold for problems which have high
initial accuracy and where state changes are infrequent. 
\end{itemize}
The ICA system was designed and tested on the task of part-of-speech
tagging, achieving an impressive reduction in training time while
suffering only a small decrease in accuracy. The experiments presented
in Section \ref{section:experiments} include ICA in the training
time and performance comparisons\footnote{%
The algorithm was implemented by the the authors, following the description
in \newcite{hepple00:tbl}.
}.

\subsubsection{Other Approaches}

\newcite{samuel98:lazy_tbl} proposed a Monte Carlo approach to transformation-based
learning, in which only a fraction of the possible rules are randomly
selected for estimation at each iteration. The \( \mu  \)-TBL system
described in \newcite{lager99:tbl} attempts to cut down on training
time with a more efficient Prolog implementation and an implementation
of ``lazy'' learning. The application of a transformation-based learning
can be considerably sped-up if the rules are compiled in a finite-state
transducer, as described in \newcite{roche94}.
 
\section{The Algorithm}

\label{section:description}

The approach presented here builds on the same foundation as the one
in \cite{ramshaw94}: instead of regenerating the rules each time,
they are stored into memory, together with the two values \( good\left( r\right)  \)
and \( bad\left( r\right)  \).

The following notations will be used throughout this section:

\begin{itemize}
\item \( G\left( r\right) =\{s\in \mathcal{S}|p_{r}(s)=true\textrm{ and }C[s]\neq t_{r}\textrm{ and }t_{r}=T[s]\} \) \,---\,
the samples on which the rule applies and changes them to the correct
classification; therefore, \( good(r)=|G(r)| \). 
\item \( B\left( r\right) =\{s\in \mathcal{S}|p_{r}(s)=true\textrm{ and }C[s]\neq t_{r}\textrm{ and }C[s]=T[s]\} \) \,---\,
the samples on which the rule applies and changes the classification
from correct to incorrect; similarly, \( bad(r)=|B(r)| \). 
\end{itemize}
Given a newly learned rule \( b \) that is to be applied to \( \mathcal{S} \),
the goal is to identify the rules \( r \) for which at least one
of the sets \( G\left( r\right) ,B\left( r\right)  \) is modified
by the application of rule \( b \). Obviously, if both sets are not
modified when applying rule \( b \), then the value of the objective
function for rule \( r \) remains unchanged.

The presentation is complicated by the fact that, in many NLP tasks,
the samples are not independent. For instance, in POS tagging, a sample
is dependent on the classification of the preceding and succeeding
2 samples (this assumes that there exists a natural ordering of the
samples in \( \mathcal{S} \)). Let \( V\left( s\right)  \) denote
the {}``vicinity{}'' of a sample \,---\, the set of samples on whose
classification the sample \( s \) might depend on (for consistency,
\( s\in V(s) \)); if samples are independent, then \( V\left( s\right) =\left\{ s\right\}  \).

\subsection{Generating the Rules}

Let \( s \) be a sample on which the best rule \( b \) applies (i.e.
\( \left[ b\left( s\right) \right] \neq C\left[ s\right]  \)). We
need to identify the rules \( r \) that are influenced by the change
\( s\rightarrow b\left( s\right)  \). Let \( r \) be such a rule.
\( f\left( r\right)  \) needs to be updated if and only if there
exists at least one sample \( s' \) such that \begin{eqnarray}
s'\in G\left( r\right) \textrm{ and }b\left( s'\right) \notin G\left( r\right) \textrm{ or} &  & \label{case1.1} \\
s'\in B\left( r\right) \textrm{ and }b\left( s'\right) \notin B\left( r\right) \textrm{ or} &  & \label{case1.2} \\
s'\notin G\left( r\right) \textrm{ and }b\left( s'\right) \in G\left( r\right) \textrm{ or} &  & \label{case1.3} \\
s'\notin B\left( r\right) \textrm{ and }b\left( s'\right) \in B\left( r\right) \, \, \, \,  & \label{case1.4} 
\end{eqnarray}
 Each of the above conditions corresponds to a specific update of
the \( good\left( r\right)  \) or \( bad\left( r\right)  \) counts.
We will discuss how rules which should get their \( good \) or \( bad \)
counts decremented (subcases (\ref{case1.1}) and (\ref{case1.2}))
can be generated, the other two being derived in a very similar fashion.

The key observation behind the proposed algorithm is: when investigating
the effects of applying the rule \( b \) to sample \( s \), only
samples \( s' \) in the set \( V\left( s\right)  \) need to be checked.
Any sample \( s' \) that is not in the set \[
\operatornamewithlimits {\bigcup }_{\left\{ s|b\textrm{ changes}\, s\right\} }V\left( s\right) \]
 can be ignored since \( s'=b(s') \).

Let \( s'\in V\left( s\right)  \) be a sample in the vicinity of
\( s \). There are 2 cases to be examined \,---\, one in which \( b \)
applies to \( s' \) and one in which \( b \) does not:

{\raggedright \vspace*{1.3mm}\textbf{Case I}: \( c\left( s'\right) =c\left( b\left( s'\right) \right)  \)
(\( b \) does not modify the classification of sample \( s' \)).
We note that the condition \[
s'\in G\left( r\right) \textrm{ and }b\left( s'\right) \notin G\left( r\right) \]
 is equivalent to\begin{equation}
\label{case11}
\begin{array}{l}
p_{r}\left( s'\right) =true\, \wedge \, C\left[ s'\right] \neq t_{r}\, \wedge \\
\, \, \, \, \, t_{r}=T\left[ s'\right] \, \wedge \, p_{r}\left( b\left( s'\right) \right) =false
\end{array}
\end{equation}
 and the formula \[
s'\in B\left( r\right) \textrm{ and }b\left( s'\right) \notin B\left( r\right) \]
 is equivalent to \begin{equation}
\label{case12}
\begin{array}{r}
p_{r}\left( s'\right) =true\wedge C\left[ s'\right] \neq t_{r}\wedge \, \, \, \, \, \, \, \, \, \, \, \, \, \, \, \, \, \, \, \, \\
C\left[ s'\right] =T\left[ s'\right] \wedge p_{r}\left( b\left( s'\right) \right) =false
\end{array}
\end{equation}
 (for the full details of the derivation, inferred from the definition
of \( G\left( r\right)  \) and \( B\left( r\right)  \), please refer
to \newcite{hans2000}).\par}

These formulae offer us a method of generating the rules \( r \)
which are influenced by the modification \( s'\rightarrow b\left( s'\right)  \):

\begin{enumerate}
\item Generate all predicates \( p \) (using the predicate templates) that
are true on the sample \( s' \). 
\item If \( C\left[ s'\right] \neq T\left[ s'\right]  \) then

\begin{enumerate}
\item \vspace*{-2mm}If \( p\left( b\left( s'\right) \right) =false \) then
decrease \( good\left( r\right)  \), where \( r \) is the rule created
with predicate \( p \) s.t. target \( T\left[ s'\right]  \); 
\end{enumerate}
\item \vspace*{-2mm}Else

\begin{enumerate}
\item \vspace*{-2mm}If \( p\left( b\left( s'\right) \right) =false \) then
for all the rules \( r \) whose predicate is \( p \)\footnote{%
This can be done efficiently with an appropriate data structure -
for example, using a double hash. 
} and \( t_{r}\neq C\left[ s'\right]  \) decrease \( bad\left( r\right)  \); 
\end{enumerate}
\end{enumerate}
The algorithm for generating the rules \( r \) that need their \( good \)
counts (formula (\pageref{case1.3})) or \( bad \) counts (formula
(\ref{case1.4})) increased can be obtained from the formulae (\ref{case1.1})
(respectively (\ref{case1.2})), by switching the states \( s' \)
and \( b\left( s'\right)  \), and making sure to add all the new
possible rules that might be generated (only for (\ref{case1.3})).

\textbf{Case II}: \( C\left[ s'\right] \neq C\left[ b\left( s'\right) \right]  \)
(\( b \) does change the classification of sample \( s' \)). In
this case, the formula (\ref{case11}) is transformed into:\hspace*{-15mm}\begin{equation}
\label{case21}
\begin{array}{r}
p_{r}\left( s'\right) =true\textrm{ }\wedge \textrm{ C}\left[ s'\right] \neq t_{r}\textrm{ }\wedge \textrm{ }t_{r}=T\left[ s'\right] \textrm{ }\wedge \\
\left( p_{r}\left( b\left( s'\right) \right) =false\vee t_{r}=C\left[ b\left( s'\right) \right] \right) 
\end{array}
\end{equation}
 (again, the full derivation is presented in \newcite{hans2000}).
The case of (\ref{case1.2}), however, is much simpler. It is easy
to notice that \( C\left[ s'\right] \neq C\left[ b\left( s'\right) \right]  \)
and \( s'\in B\left( r\right)  \) implies that \( b\left( s'\right) \notin B\left( r\right)  \);
indeed, a necessary condition for a sample \( s' \) to be in a set
\( B\left( r\right)  \) is that \( s' \) is classified correctly,
\( C\left[ s'\right] =T\left[ s'\right]  \). Since \( T\left[ s'\right] \neq C\left[ b\left( s'\right) \right]  \),
results \( C\left[ b\left( s'\right) \right] \neq T\left[ s'\right]  \)
and therefore \( b\left( s'\right) \notin B\left( r\right)  \). Condition
(\pageref{case1.2}) is, therefore, equivalent to \begin{equation}
\label{case22}
p_{r}\left( s'\right) =true\textrm{ }\wedge \textrm{ C}\left[ s'\right] \neq t_{r}\textrm{ }\wedge \textrm{ C}\left[ s'\right] =T\left[ s'\right] 
\end{equation}
 The algorithm is modified by replacing the test \( p\left( b\left( s'\right) \right) =false \)
with the test \( p_{r}\left( b\left( s'\right) \right) =false\vee C\left[ b\left( s\right) \right] =t_{r} \)
in formula (\ref{case1.1}) and removing the test altogether for case
of (\ref{case1.2}). The formulae used to generate rules \( r \)
that might have their counts increased (equations (\ref{case1.3})
and (\ref{case1.4})) are obtained in the same fashion as in Case
I.

\subsection{The Full Picture}

\begin{figure}
\begin{tabular}{|l|}

\hline

\parbox[t]{3in}{

\setlength{\labelsep}{0.5mm}

\setlength{\leftmargin}{0in}

\setlength{\labelwidth}{0.5mm}

{\footnotesize For all samples \( s \) that satisfy \( C\left[ s\right] \neq T\left[ s\right]  \),
generate all rules \( r \) that correct the classification of \( s \);
increase \( good\left( r\right)  \).}{\footnotesize \par}

{\footnotesize For all samples \( s \) that satisfy \( C\left[ s\right] =T\left[ s\right]  \)
generate all predicates \( p \) s.t. \( p\left( s\right) =true \);
for each rule \( r \) s.t. \( p_{r}=p \) and \( t_{r}\neq C\left[ s\right]  \)
increase \( bad\left( r\right)  \).}{\footnotesize \par}

{\footnotesize 1: Find the rule \( b=\arg \max _{r\in \mathcal{R}}f\left( r\right)  \).}{\footnotesize \par}

{\footnotesize If (\( f\left( b\right) <\textrm{Threshold} \) or
corpus learned to completion) then quit.}{\footnotesize \par}

{\footnotesize For each predicate \( p \), let \( \mathcal{R}\left( p\right)  \)
be the rules whose predicate is \( p \) (\( p_{r}=r \)).}{\footnotesize \par}

{\footnotesize For each samples \( s,s' \) s.t. \( C\left[ s\right] \neq C\left[ b\left( s\right) \right]  \)
and \( s'\in V\left( s\right)  \):}{\footnotesize \par}

{\footnotesize If \( C\left[ s'\right] =C\left[ b\left( s'\right) \right]  \)
then}{\footnotesize \par}

\begin{itemize}
\item {\footnotesize \vspace*{-2.5mm}for each predicate \( p \) s.t. \( p\left( s'\right) =true \)}{\footnotesize \par}

\begin{itemize}
\item {\footnotesize \vspace*{-2.5mm}If \( C\left[ s'\right] \neq T\left[ s'\right]  \)
then}{\footnotesize \par}

\begin{itemize}
\item {\footnotesize \vspace*{-1.5mm}If \( p\left( b\left( s'\right) \right) =false \)
then decrease \( good\left( r\right)  \), where \( r=\left[ p,T\left[ s'\right] \right]  \),
the rule created with predicate \( p \) and target \( T\left[ s'\right]  \); }{\footnotesize \par}
\end{itemize}
\item {\footnotesize \vspace*{-1.5mm}Else}{\footnotesize \par}

\begin{itemize}
\item {\footnotesize \vspace*{-1.5mm}If \( p\left( b\left( s'\right) \right) =false \)
then for all the rules \( r\in \mathcal{R}\left( p\right)  \) s.t.
\( t_{r}\neq C\left[ s'\right]  \) decrease \( bad\left( r\right)  \); }{\footnotesize \par}
\end{itemize}
\end{itemize}
\item {\footnotesize \vspace*{-2.5mm}for each predicate \( p \) s.t. \( p\left( b\left( s'\right) \right) =true \)}{\footnotesize \par}

\begin{itemize}
\item {\footnotesize \vspace*{-2mm}If \( C\left[ b\left( s'\right) \right] \neq T\left[ s'\right]  \)
then}{\footnotesize \par}

\begin{itemize}
\item {\footnotesize \vspace*{-2mm}If \( p\left( s'\right) =false \) then
increase \( good\left( r\right)  \), where \( r=\left[ p,T\left[ s'\right] \right]  \); }{\footnotesize \par}
\end{itemize}
\item {\footnotesize \vspace*{-1.5mm}Else}{\footnotesize \par}

\begin{itemize}
\item {\footnotesize \vspace*{-1.5mm}If \( p\left( s'\right) =false \)
then for all rules \( r\in \mathcal{R}\left( p\right)  \) s.t. \( t_{r}\neq C\left[ b\left( s'\right) \right]  \)
increase \( bad\left( r\right)  \); }{\footnotesize \par}
\end{itemize}
\end{itemize}
\end{itemize}
{\footnotesize \vspace*{-3mm}Else}{\footnotesize \par}

\begin{itemize}
\item {\footnotesize \vspace*{-2.5mm}for each predicate \( p \) s.t. \( p\left( s'\right) =true \)}{\footnotesize \par}

\begin{itemize}
\item {\footnotesize \vspace*{-2mm}If \( C\left[ s'\right] \neq T\left[ s'\right]  \)
then}{\footnotesize \par}

\begin{itemize}
\item {\footnotesize \vspace*{-1mm}If \( p\left( b\left( s'\right) \right) =false\vee C\left[ b\left( s'\right) \right] =t_{r} \)
then decrease \( good\left( r\right)  \), where \( r=\left[ p,T\left[ s'\right] \right]  \); }{\footnotesize \par}
\end{itemize}
\item {\footnotesize \vspace*{-1mm}Else}{\footnotesize \par}

\begin{itemize}
\item {\footnotesize \vspace*{-1mm}For all the rules \( r\in \mathcal{R} \)\( \left( p\right)  \)
s.t. \( t_{r}\neq C\left[ s'\right]  \) decrease \( bad\left( r\right)  \); }{\footnotesize \par}
\end{itemize}
\end{itemize}
\item {\footnotesize \vspace*{-2.5mm}for each predicate \( p \) s.t. \( p\left( b\left( s'\right) \right) =true \)}{\footnotesize \par}

\begin{itemize}
\item {\footnotesize \vspace*{-2.5mm}If \( C\left[ b\left( s'\right) \right] \neq T\left[ s'\right]  \)
then}{\footnotesize \par}

\begin{itemize}
\item {\footnotesize \vspace*{-1mm}If \( p\left( s'\right) =false\vee C\left[ s'\right] =t_{r} \)
then increase \( good\left( r\right)  \), where \( r=\left[ p,T\left[ s'\right] \right]  \); }{\footnotesize \par}
\end{itemize}
\item {\footnotesize \vspace*{-1mm}Else}{\footnotesize \par}

\begin{itemize}
\item {\footnotesize \vspace*{-1mm}For all rules \( r\in \mathcal{R}\left( p\right)  \)
s.t. \( t_{r}\neq C\left[ b\left( s'\right) \right]  \) increase
\( bad\left( r\right)  \); }{\footnotesize \par}
\end{itemize}
\end{itemize}
\end{itemize}
{\footnotesize \vspace*{-3mm}Repeat from step 1:}{\footnotesize \par}

}\\

\hline

\end{tabular}

\vspace{0pt}
\caption{FastTBL Algorithm \label{fast-tbl1}}
\end{figure}
At every point in the algorithm, we assumed that all the rules that
have at least some positive outcome (\( good\left( r\right) >0 \))
are stored, and their score computed. Therefore, at the beginning
of the algorithm, all the rules that correct at least one wrong classification
need to be generated. The bad counts for these rules are then computed
by generation as well: in every position that has the correct classification,
the rules that change the classification are generated, as in Case
\ref{case1.4}, and their bad counts are incremented. The entire FastTBL
algorithm is presented in Figure \ref{fast-tbl1}. Note that, when
the bad counts are computed, only rules that already have positive
good counts are selected for evaluation. This prevents the generation
of useless rules and saves computational time. 

The number of examined rules is kept close to the minimum. Because
of the way the rules are generated, most of them need to modify either
one of their counts. Some additional space (besides the one needed
to represent the rules) is necessary for representing the rules in
a predicate hash \,---\, in order to have a straightforward access
to all rules that have a given predicate; this amount is considerably
smaller than the one used to represent the rules. For example, in
the case of text chunking task described in section \ref{section:experiments}, only
approximately 30Mb additional memory is required, while the approach
of \newcite{ramshaw94} would require approximately 450Mb.

\subsection{Behavior of the Algorithm}

\label{subsection:tbl_behavior}

As mentioned before, the original algorithm has a number of deficiencies
that cause it to run slowly. Among them is the drastic slowdown in
rule learning as the scores of the rules decrease. When the best rule
has a high score, which places it outside the tail of the score distribution,
the rules in the tail will be skipped when the bad counts are calculated,
since their good counts are small enough to cause them to be discarded.
However, when the best rule is in the tail, many other rules with
similar scores can no longer be discarded and their bad counts need
to be computed, leading to a progressively longer running time per
iteration.

Our algorithm does not suffer from the same problem, because the counts
are updated (rather than recomputed) at each iteration, and only for
the samples that were affected by the application of the latest rule
learned. Since the number of affected samples decreases as learning
progresses, our algorithm actually \textit{speeds up} considerably
towards the end of the training phase. Considering that the number
of low-score rules is a considerably higher than the number of high-score
rules, this leads to a dramatic reduction in the overall running time.

This has repercussions on the scalability of the algorithm relative
to training data size. Since enlarging the training data size results
in a longer score distribution tail, our algorithm is expected to
achieve an even more substantial relative running time improvement
over the original algorithm. Section \ref{section:experiments} presents
experimental results that validate the superior scalability of the
FastTBL algorithm.

\section{Experiments\label{section:experiments}}

Since the goal of this paper is to compare and contrast system training
time and performance, extra measures were taken to ensure fairness
in the comparisons. To minimize implementation differences, all the
code was written in C++ and classes were shared among the systems
whenever possible. For each task, the same training set was provided
to each system, and the set of possible rule templates was kept the
same. Furthermore, extra care was taken to run all comparable experiments
on the same machine and under the same memory and processor load conditions. 

To provide a broad comparison between the systems, three NLP tasks
with different properties were chosen as the experimental domains.
The first task, part-of-speech tagging, is one where the commitment
assumption seems intuitively valid and the samples are not independent.
The second task, prepositional phrase attachment, has examples which
are independent from each other. The last task is text chunking, where
both independence and commitment assumptions do not seem to be valid.
A more detailed description of each task, data and the system parameters
are presented in the following subsections.

Four algorithms are compared during the following experiments:

\begin{itemize}
\item \vspace*{-1mm}The regular TBL, as described in section \ref{section:tbl};
\item \vspace*{-1mm}An improved version of TBL, which makes extensive use
of indexes to speed up the rules' update;
\item \vspace*{-1mm}The FastTBL algorithm;
\item \vspace*{-1mm}The ICA algorithm \cite{hepple00:tbl}.
\end{itemize}

\subsection{Part-of-Speech Tagging}

\begin{table*}
{\centering \begin{tabular}{|c|c|c|c|c|c|}
\hline 
&
\emph{\small Brill's tagger}&
\textit{\small Regular TBL}&
 \emph{}\textit{\small Indexed} \emph{\small TBL}&
 \emph{}\emph{\small FastTBL}&
 \emph{}\emph{\small ICA (Hepple)}\\
\hline
\textit{\small Accuracy}&
\( 96.61\% \)&
\( 96.61\% \)&
\( 96.61\% \) &
\( 96.61\% \)&
\( 96.23\% \)\\
\hline
\emph{\small Running time}&
 {\small 5879 mins, 46 secs}&
{\small 2286 mins, 21 secs}&
\textit{\emph{\small 420 mins, 7 secs}}&
{\small 17 mins, 21 secs}&
{\small 6 mins, 13 secs}\\
\hline
\emph{\small Time ratio}&
\( 0.4 \)&
\( 1.0 \)&
\( 5.4 \)&
\( 131.7 \)&
\( 367.8 \)\\
\hline
\end{tabular}\par}

\caption{POS tagging: Evaluation and Running Times\label{pos-results}}
\end{table*}

The goal of this task is to assign to each word in the given sentence
a tag corresponding to its part of speech. A multitude of approaches
have been proposed to solve this problem, including transformation-based
learning, Maximum Entropy models, Hidden Markov models and memory-based
approaches.

The data used in the experiment was selected from the Penn Treebank
Wall Street Journal, and is the same used by \newcite{brill98}. The
training set contained approximately 1M words and the test set approximately
200k words.

Table \ref{pos-results} presents the results of the experiment\footnote{%
The time shown is the combined running time for both the lexical tagger
and the contextual tagger.
}. All the algorithms were trained until a rule with a score of 2 was
reached. The FastTBL algorithm performs very similarly to the regular
TBL, while running in an order of magnitude faster. The two assumptions
made by the ICA algorithm result in considerably less training time,
but the performance is also degraded (the difference in performance
is statistically significant, as determined by a signed test, at a
significance level of \( 0.001 \)). Also present in Table \ref{pos-results}
are the results of training Brill's tagger on the same data. The results
of this tagger are presented to provide a performance comparison with
a widely used tagger. Also worth mentioning is that the tagger achieved
an accuracy of \( 96.76\% \) when trained on the entire data\footnote{%
We followed the setup from Brill's tagger: the contextual tagger is
trained only on half of the training data. The training time on the
entire data was approximately 51 minutes.
}; a Maximum Entropy tagger \cite{ratnaparkhi96} achieves \( 96.83\% \)
accuracy with the same training data/test data.

\subsection{Prepositional Phrase Attachment}

\begin{table*}
{\centering \begin{tabular}{|c|c|c|c|c|}
\hline 
&
\centering  \emph{}\textit{\small Regular TBL}&
\textit{\small Indexed TBL}&
 \textit{\small Fast TBL}&
 \textit{\small ICA (Hepple}\textit{\emph{\small )}}\\
\hline
\textit{\small Accuracy}&
\( 81.0\% \)&
\( 81.0\% \)&
\( 81.0\% \)&
 \( 77.8\% \)\\
\hline
\textit{\small Running time}&
\textit{\emph{\small 190 mins, 19 secs}}&
\textit{\emph{\small 65 mins, 50 secs}}&
 \textit{\emph{\small 14 mins, 38 secs}}&
 \textit{\emph{\small 4 mins, 1 sec}} \\
\hline
\textit{\small Time Ratio}&
\( 1.0 \)&
\( 2.9 \)&
\( 13 \)&
\( 47.4 \)\\
\hline
\end{tabular}\par}

\caption{PP Attachment:Evaluation and Running Times\label{pp_results}}
\end{table*}

Prepositional phrase attachment is the task of deciding the point
of attachment for a given prepositional phrase (PP). As an example,
consider the following two sentences:

\begin{enumerate}
\item \vspace*{-2mm}\label{pp:ex1}I washed the shirt with soap and water. 
\item \vspace*{-1mm}\label{pp:ex2}I washed the shirt with pockets. 
\end{enumerate}
\vspace*{-2mm}In Sentence \ref{pp:ex1}, the PP \textit{{}``with
soap and water{}''} describes the act of washing the shirt. In Sentence
\ref{pp:ex2}, however, the PP \textit{{}``with pockets{}''} is
a description for the shirt that was washed.

Most previous work has concentrated on situations which are of the
form \textit{VP NP1 P NP2}. The problem is cast as a classification
task, and the sentence is reduced to a 4-tuple containing the preposition
and the non-inflected base forms of the head words of the verb phrase
\textit{VP} and the two noun phrases \textit{NP1} and \textit{NP2}.
For example, the tuple corresponding to the two above sentences would
be:

\begin{enumerate}
\item \vspace*{-2mm}wash shirt with soap
\item \vspace*{-1mm}wash shirt with pocket 
\end{enumerate}
\vspace*{-2mm}Many approaches to solving this this problem have been
proposed, most of them using standard machine learning techniques,
including transformation-based learning, decision trees, maximum entropy
and backoff estimation. The transformation-based learning system was
originally developed by \newcite{brill94:pp-attach}.

The data used in the experiment consists of approximately 13,000 \textit{\emph{quadruples
(}}\textit{VP} \textit{\emph{}}\textit{NP1 P NP2}\textit{\emph{)}}
extracted from Penn Treebank parses. The set is split into a test
set of 500 samples and a training set of 12,500 samples. The templates
used to generate rules are similar to the ones used by \newcite{brill94:pp-attach}
and some include WordNet features. All the systems were trained until
no more rules could be learned.

Table \ref{pp_results} shows the results of the experiments. Again,
the ICA algorithm learns the rules very fast, but has a slightly lower
performance than the other two TBL systems. Since the samples are
inherently independent, there is no performance loss because of the
independence assumption; therefore the performance penalty has to
come from the commitment assumption. The Fast TBL algorithm runs,
again, in a order of magnitude faster than the original TBL while
preserving the performance; the time ratio is only 13 in this case
due to the small training size (only 13000 samples).

\subsection{Text Chunking}

\begin{table*}
{\centering \begin{tabular}{|c|c|c|c|c|}
\hline 
&
\textit{\small Regular TBL}&
 \textit{\small Indexed TBL}&
 \textit{\small Fast TBL}&
 \textit{\small ICA (Hepple)}\\
\hline
\textit{\small F-measure}&
92.30&
 92.30&
 92.30&
 86.20\\
\hline
\textit{\small Running Time} \emph{}&
{\small 19211 mins, 40 secs}&
 \textit{\emph{\small 2056 mins, 4secs}}&
 \textit{\emph{\small 137 mins, 57 secs}}&
 \textit{\emph{\small 12 mins, 40 secs}}  \\
\hline 
\textit{\small Time Ratio}&
\( 1.0 \)&
\( 9.3 \)&
\( 139.2 \)&
\( 1516.7 \)\\
\hline
\end{tabular}\par}

\caption{Text Chunking: Evaluation and Running Times\label{text-chunking-results}}
\end{table*}

Text chunking is a subproblem of syntactic parsing, or sentence diagramming.
Syntactic parsing attempts to construct a parse tree from a sentence
by identifying all phrasal constituents and their attachment points.
Text chunking simplifies the task by dividing the sentence into non-overlapping
phrases, where each word belongs to the lowest phrasal constituent
that dominates it. The following example shows a sentence with text
chunks and part-of-speech tags: \vspace*{-0.1cm}

\begin{quote}
\textbf{{[}NP} A.P.\( _{NNP} \) Green\( _{NNP} \) \textbf{{]} {[}ADVP}
currently\( _{RB} \) \textbf{{]} {[}VP} has \textbf{{]} {[}NP} 2,664,098\( _{CD} \)
shares\( _{NNS} \)\textbf{{]} {[}ADJP} outstanding\( _{JJ} \) \textbf{{]}}
.\vspace*{-0.1cm}
\end{quote}
The problem can be transformed into a classification task. Following
Ramshaw \& Marcus' \shortcite{ramshaw99:basenp} work in base noun
phrase chunking, each word is assigned a chunk tag corresponding to
the phrase to which it belongs . The following table shows the above
sentence with the assigned chunk tags:

{\centering \begin{tabular}{|c|c|c|}
\hline 
{\small Word }&
 {\small POS tag} &
 {\small Chunk Tag}\\
\hline
{\small A.P.} &
 {\small NNP }&
 {\small B-NP}\\
 {\small Green }&
 {\small NNP }&
 {\small I-NP}\\
 {\small currently }&
 {\small RB }&
 {\small B-ADVP}\\
 {\small has} &
 {\small VBZ} &
 {\small B-VP}\\
 {\small 2,664,098} &
 {\small CD} &
 {\small B-NP}\\
 {\small shares }&
 {\small NNS} &
 {\small I-NP}\\
 {\small outstanding }&
 {\small JJ }&
 {\small B-ADJP}\\
 . &
 . &
 {\small O}  \\
\hline
\end{tabular}\par}

The data used in this experiment is the CoNLL-2000 phrase chunking
corpus \cite{tjong00:conll_shared}. The training corpus consists
of sections 15-18 of the Penn Treebank \cite{penntreebank}; section
20 was used as the test set. The chunk tags are derived from the parse
tree constituents, and the part-of-speech tags were generated by Brill's
tagger \cite{brill95:tagging}. All the systems are trained to completion
(until all the rules are learned).

Table \ref{text-chunking-results} shows the results of the text chunking
experiments. The performance of the FastTBL algorithm is the same
as of regular TBL's, and runs in an order of magnitude faster. The
ICA algorithm again runs considerably faster, but at a cost of a significant
performance hit. There are at least 2 reasons that contribute to this
behavior:

\begin{enumerate}
\item \vspace*{-2.4mm}The initial state has a lower performance than the
one in tagging; therefore the independence assumption might not hold.
25\% of the samples are changed by at least one rule, as opposed to
POS tagging, where only 2.5\% of the samples are changed by a rule. 
\item \vspace*{-1mm}The commitment assumption might also not hold. For this
task, 20\% of the samples that were modified by a rule are also changed
again by another one. 
\end{enumerate}

\subsection{Training Data Size Scalability}

\begin{figure*}
{\centering \hspace*{-5mm}\vspace*{-5mm}\begin{tabular}{lc}
\subfigure[Running Time versus Training Data Size \label{scalability}]{\resizebox*{0.5\textwidth}{5cm}{\includegraphics{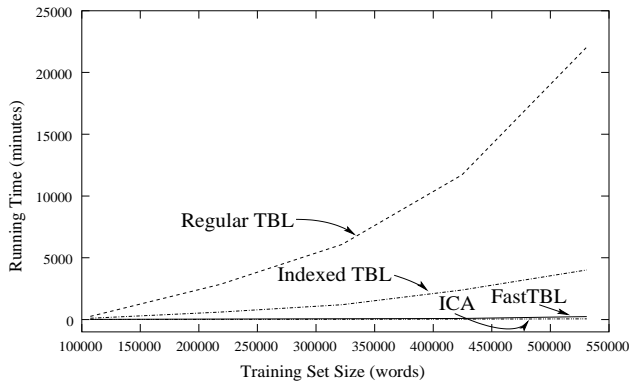}}} &
\subfigure[\label{iteration-plot} Running Time versus Iteration Number]{\resizebox*{0.5\textwidth}{5cm}{\includegraphics{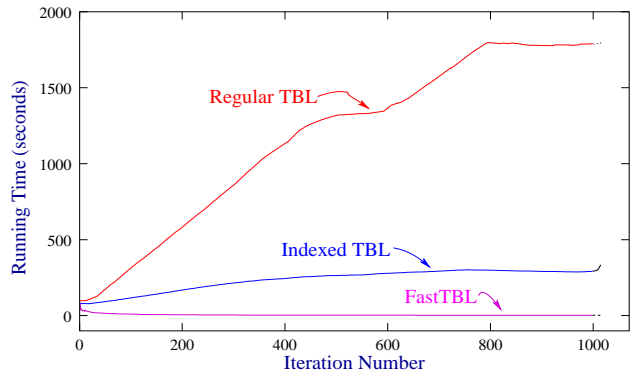}}} \\
\end{tabular}\par}

\caption{\vspace*{-10mm}Algorithm Scalability }
\end{figure*}
A question usually asked about a machine learning algorithm is how
well it adapts to larger amounts of training data. Since the performance
of the Fast TBL algorithm is identical to that of regular TBL, the
issue of interest is the dependency between the running time of the
algorithm and the amount of training data.

The experiment was performed with the part-of-speech data set. The
four algorithms were trained on training sets of different sizes;
training times were recorded and averaged over 4 trials. The results
are presented in Figure \ref{scalability}. It is obvious that the
Fast TBL algorithm is much more scalable than the regular TBL \,---\,
displaying a linear dependency on the amount of training data, while
the regular TBL has an almost quadratic dependency. The explanation
for this behavior has been given in Section \ref{subsection:tbl_behavior}.

Figure \ref{iteration-plot} shows the time spent at each iteration
versus the iteration number, for the original TBL and fast TBL systems.
It can be observed that the time taken per iteration increases dramatically
with the iteration number for the regular TBL, while for the FastTBL,
the situation is reversed. The consequence is that, once a certain
threshold has been reached, the incremental time needed to train the
FastTBL system to completion is negligible.

\section{Conclusions\label{section:conclusions}}

We have presented in this paper a new and improved method of computing
the objective function for transformation-based learning. This method
allows a transformation-based algorithm to train an observed 13 to
139 times faster than the original one, while preserving the final
performance of the algorithm. The method was tested in three different
domains, each one having different characteristics: part-of-speech
tagging, prepositional phrase attachment and text chunking. The results
obtained indicate that the algorithmic improvement generated by our
method is not linked to a particular task, but extends to any classification
task where transformation-based learning can be applied. Furthermore,
our algorithm scales better with training data size; therefore the
relative speed-up obtained will increase when more samples are available
for training, making the procedure a good candidate for large corpora
tasks.

The increased speed of the Fast TBL algorithm also enables its usage
in higher level machine learning algorithms, such as adaptive boosting,
model combination and active learning. Recent work \cite{florian00:tbldt}
has shown how a TBL framework can be adapted to generate confidences
on the output, and our algorithm is compatible with that framework.
The stability, resistance to overtraining, the existence of probability
estimates and, now, reasonable speed make TBL an excellent candidate
for solving classification tasks in general.

\section{Acknowledgements}

The authors would like to thank David Yarowsky for his advice and
guidance, Eric Brill and John C. Henderson for discussions on the
initial ideas of the material presented in the paper, and the anonymous
reviewers for useful suggestions, observations and connections with
other published material. The work presented here was supported by
NSF grants IRI-9502312 and IRI-9618874.

\bibliographystyle{acl}
\bibliography{tbl-speedup}

\begin{thebibliography}{}

\bibitem[\protect\citename{Brill and Resnik}1994]{brill94:pp-attach}
E.~Brill and P.~Resnik.
\newblock 1994.
\newblock A rule-based approach to prepositional phrase attachment
  disambiguation.
\newblock In {\em Proceedings of the Fifteenth International Conference on
  Computational Linguistics (COLING-1994)}, pages 1198--1204, Kyoto.

\bibitem[\protect\citename{Brill and Wu}1998]{brill98}
E.~Brill and J.~Wu.
\newblock 1998.
\newblock Classifier combination for improved lexical disambiguation.
\newblock {\em Proceedings of COLING-ACL'98}, pages 191--195, August.

\bibitem[\protect\citename{Brill}1995]{brill95:tagging}
E.~Brill.
\newblock 1995.
\newblock Transformation-based error-driven learning and natural language
  processing: A case study in part of speech tagging.
\newblock {\em Computational Linguistics}, 21(4):543--565.

\bibitem[\protect\citename{Brill}1996]{brill96:transformation_parsing}
E.~Brill, 1996.
\newblock {\em Recent Advances in Parsing Technology}, chapter Learning to
  Parse with Transformations.
\newblock Kluwer.

\bibitem[\protect\citename{Day \bgroup et al.\egroup }1997]{day97:alembic}
D.~Day, J.~Aberdeen, L.~Hirschman, R.~Kozierok, P.~Robinson, and M.~Vilain.
\newblock 1997.
\newblock Mixed-initiative development of language processing systems.
\newblock In {\em Fifth Conference on Applied Natural Language Processing},
  pages 348--355. Association for Computational Linguistics, March.

\bibitem[\protect\citename{Florian and Ngai}2001]{hans2000}
R.~Florian and G.~Ngai.
\newblock 2001.
\newblock Transformation-based learning in the fast lane.
\newblock Technical report, Johns Hopkins University, Computer Science
  Department.

\bibitem[\protect\citename{Florian \bgroup et al.\egroup
  }2000]{florian00:tbldt}
R.~Florian, J.C. Henderson, and G.~Ngai.
\newblock 2000.
\newblock Coaxing confidence from an old friend: Probabilistic classifications
  from transformation rule lists.
\newblock In {\em Proceedings of SIGDAT-EMNLP 2000}, pages 26--34, Hong Kong,
  October.

\bibitem[\protect\citename{Hepple}2000]{hepple00:tbl}
M.~Hepple.
\newblock 2000.
\newblock Independence and commitment: Assumptions for rapid training and
  execution of rule-based pos taggers.
\newblock In {\em Proceedings of the 38th Annual Meeting of the ACL}, pages
  278--285, Hong Kong, October.

\bibitem[\protect\citename{Lager}1999]{lager99:tbl}
T.~Lager.
\newblock 1999.
\newblock The $\mu$-tbl system: Logic programming tools for
  transformation-based learning.
\newblock In {\em Proceedings of the 3rd International Workshop on
  Computational Natural Language Learning}, Bergen.

\bibitem[\protect\citename{Mangu and Brill}1997]{mangu97:cssc}
L.~Mangu and E.~Brill.
\newblock 1997.
\newblock Automatic rule acquisition for spelling correction.
\newblock In {\em Proceedings of the Fourteenth International Conference on
  Machine Learning}, pages 734--741, Nashville, Tennessee.

\bibitem[\protect\citename{Marcus \bgroup et al.\egroup }1993]{penntreebank}
M.~P. Marcus, B.~Santorini, and M.~A. Marcinkiewicz.
\newblock 1993.
\newblock Building a large annotated corpus of english: The {Penn Treebank}.
\newblock {\em Computational Linguistics}, 19(2):313--330.

\bibitem[\protect\citename{Ramshaw and Marcus}1994]{ramshaw94}
L.~Ramshaw and M.~Marcus.
\newblock 1994.
\newblock Exploring the statistical derivation of transformational rule
  sequences for part-of-speech tagging.
\newblock In {\em The Balancing Act: Proceedings of the ACL Workshop on
  Combining Symbolic and Statistical Approaches to Language}, pages 128--135,
  New Mexico State University, July.

\bibitem[\protect\citename{Ramshaw and Marcus}1999]{ramshaw99:basenp}
L.~Ramshaw and M.~Marcus, 1999.
\newblock {\em Natural Language Processing Using Very Large Corpora}, chapter
  Text Chunking Using Transformation-based Learning, pages 157--176.
\newblock Kluwer.

\bibitem[\protect\citename{Ratnaparkhi}1996]{ratnaparkhi96}
A.~Ratnaparkhi.
\newblock 1996.
\newblock A maximum entropy part-of-speech tagger.
\newblock In {\em Proceedings of the First Conference on Empirical Methods in
  NLP}, pages 133--142, Philadelphia, PA.

\bibitem[\protect\citename{Rivest}1987]{rivest87:decision_lists}
R.~Rivest.
\newblock 1987.
\newblock Learning decision lists.
\newblock {\em Machine Learning}, 2(3):229--246.

\bibitem[\protect\citename{Roche and Schabes}1995]{roche94}
E.~Roche and Y.~Schabes.
\newblock 1995.
\newblock Computational linguistics.
\newblock {\em Deterministic Part of Speech Tagging with Finite State
  Transducers}, 21(2):227--253.

\bibitem[\protect\citename{Samuel \bgroup et al.\egroup
  }1998]{samuel98:dialogact}
K.~Samuel, S.~Carberry, and K.~Vijay-Shanker.
\newblock 1998.
\newblock Dialogue act tagging with transformation-based learning.
\newblock In {\em Proceedings of the 17th International Conference on
  Computational Linguistics and the 36th Annual Meeting of the Association for
  Computational Linguistics}, pages 1150--1156, Montreal, Quebec, Canada.

\bibitem[\protect\citename{Samuel}1998]{samuel98:lazy_tbl}
K.~Samuel.
\newblock 1998.
\newblock Lazy transformation-based learning.
\newblock In {\em Proceedings of the 11th Interational Florida AI Research
  Symposium Conference}, pages 235--239, Florida, USA.

\bibitem[\protect\citename{Tjong Kim~Sang and
  Buchholz}2000]{tjong00:conll_shared}
E.~Tjong Kim~Sang and S.~Buchholz.
\newblock 2000.
\newblock Introduction to the conll-2000 shared task: Chunking.
\newblock In {\em Proceedings of CoNLL-2000 and LLL-2000}, pages 127--132,
  Lisbon, Portugal.

\end{thebibliography}

\end{document}